# Verbal Focus-of-Attention System for Learning-from-Observation

Naoki Wake, Iori Yanokura, Kazuhiro Sasabuchi, and Katsushi Ikeuchi

*Abstract*—The learning-from-observation (LfO) framework aims to map human demonstrations to a robot to reduce programming effort. To this end, an LfO system encodes a human demonstration into a series of execution units for a robot, which are referred to as task models. Although previous research has proposed successful task-model encoders, there has been little discussion on how to guide a task-model encoder in a scene with spatio-temporal noises, such as cluttered objects or unrelated human body movements. Inspired by the function of verbal instructions guiding an observer's visual attention, we propose a verbal focus-of-attention (FoA) system (i.e., spatio-temporal filters) to guide a task-model encoder. For object manipulation, the system first recognizes the name of a target object and its attributes from verbal instructions. The information serves as a where-to-look FoA filter to confine the areas in which the target object existed in the demonstration. The system then detects the timings of grasp and release that occurred in the filtered areas. The timings serve as a when-to-look FoA filter to confine the period of object manipulation. Finally, a task-model encoder recognizes the task models by employing FoA filters. We demonstrate the robustness of the verbal FoA in attenuating spatio-temporal noises by comparing it with an existing action localization network. The contributions of this study are as follows: (1) to propose a verbal FoA for LfO, (2) to design an algorithm to calculate FoA filters from verbal input, and (3) to demonstrate the effectiveness of a verbal FoA in localizing an action by comparing it with a state-of-the-art vision system.

## I. INTRODUCTION

In several societies, there is a rising demand for service robots who perform household operations for elderly people. A survey of older adults (65–93 years old) revealed that one required operation is manipulating objects regardless of purpose or context [1]. To this end, one possible approach is to prepare a robot teaching system that can perform various template manipulation operations that are optimized for individual homes through on-site demonstrations by non-experts. Learning-from-observation (LfO) is a framework that can realize this concept [2], [3] (Fig. 1).

In a traditional LfO system, the task-model encoder first uses passive observation to encode human demonstrations into a series of execution units (i.e., substantiated templates) for a robot. These are referred to as *task models* [2]. A task model contains execution information regarding "what to do" and "how to do," referred to as *task* and *skill parameters*, respectively. The task-model decoder then decodes the task models with on-site visual information to calculate the appropriate motor commands for a robot in the real world.

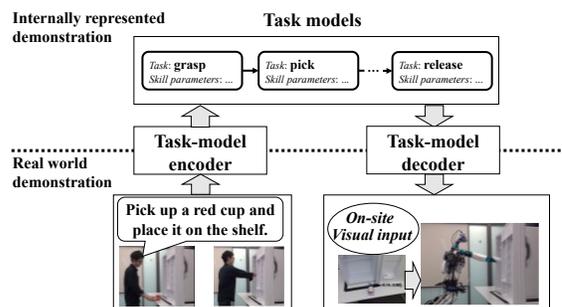

Figure 1. Learning-from-observation framework.

Although LfO research has proposed effective task-model encoders that analyze a target object during a manipulation or the trajectory of a manipulating hand [2]–[11], these studies postulate that the encoder can access the position or the time at which the manipulation occurs. In many cases, a manipulation starts with grasping a target object and finishes with releasing the object. Therefore, a challenge for deployment of an LfO system in the real world is to detect timings and locations of grasping and releasing in a typically noisy environment.

This study defines the problem to solve as "given a demonstration with noises, output the timings and locations of grasping and releasing." We define noise as any unrelated information presented to the task-model encoder. Specifically, we deal with two types of noises: spatial and temporal. We define spatial noise as unrelated objects in a demonstration. To address this noise, the task-model encoder needs to filter for an object of interest. Temporal noise is defined as unrelated human body movements before and after a manipulation operation. To address this noise, the task-model encoder needs to filter for the period of a manipulation operation.

Here, we propose an active observation system with verbally-driven spatio-temporal filters, referred to as verbal *focus-of-attention* (FoA), to guide a task-model encoder. In human-to-human teaching, verbal instructions are often used in conjunction with visual instructions to efficiently guide an observer's visual attention [12]–[14]. Inspired by the function of verbal instructions, we propose the application of a verbal FoA to reduce noise in a demonstration. For example, when a demonstrator says, "Open the fridge.," a system should reasonably pay attention to the fridge in the demonstration. In addition, task-related human body movements should occur while the demonstrator holds the fridge door. Once the system knows "where and when to pay attention," the task-model encoder can easily obtain skill parameters to substantiate the task models.

This study deals with the design aspects involved in producing verbal FoA, which is composed of several sub-FoA filters. The system first recognizes a task-related verb (e.g., pick) and the name of the target object with its attributes (e.g., a red cup) from verbal instructions. The information serves as a where-to-look FoA filter to confine the areas where the target object existed in the demonstration. The system then detects the timings of the grasp and release occurring in the filtered areas by analyzing the time-series distances of the object to both hands. The timings serve as a when-to-look FoA filter to confine the period over which the demonstrator manipulated the object. Finally, a task-model encoder recognizes the task models by employing FoA filters. We demonstrate the robustness of the verbal FoA by comparing it with an existing action localization network. The contributions of this study are as follows: (1) to propose a verbal FoA for LfO, (2) to design an algorithm to calculate FoA filters from verbal input, and (3) to demonstrate the effectiveness of a verbal FoA by comparing it with a state-of-the-art vision system.

## II. RELATED WORKS

### A. Robot teaching frameworks

In the context of robot teaching, popular frameworks are learning-from-demonstration (LfD) and LfO. Representative work of LfD (sometimes referred to as programming-by-demonstration) is included in high-quality surveys [15]–[17]. Although LfO and LfD are similar in that they aim to map human demonstrations to robot movements, they are based on different philosophies. LfO is based on a task-oriented programming approach [2], which operates a robot with an understanding of the purpose of tasks, whereas LfD is based on a machine-learning approach, which obtains the intermediate task representation, *the so-called policy*, through repeated observation of body trajectories during demonstrations. One drawback of LfD is that a learned policy cannot be transferred to arbitrary robot hardware. This challenge is broadly referred to as the "correspondence problem" [18]. In contrast, LfO research has avoided this problem by utilizing predefined motion templates (i.e., task model) that are not specific to hardware [2]–[11] (Fig. 1). Although LfO may be a preferable approach for teaching arbitrary robots, previous task-based LfO systems were designed in laboratory environments without noise.

Overall, our proposed verbal FoA is positioned as a solution for a task-oriented robot teaching system, LfO, to widen the applications and orient towards a practical environment with spatio-temporal noises.

### B. Focus-of-attention (FoA)

In the context of computer vision, "active recognition" is a methodology to improve image recognition by FoA [19]. Active recognition is characterized by processing the visual information in a cluttered scene by employing high-level knowledge regarding what a system needs to recognize. For example, Ikeuchi et al. proposed a task-oriented cognitive system that systematically modifies the architecture of the vision system, according to the specifications of each task [20]. The verbal FoA is a form of active recognition because it uses verbal instructions to understand the purpose of tasks. However, applications that utilize verbal FoA are, if any, limited in specific domains such as visual object search [21].

Another related research topic is to model a vision-based FoA, which is called "saliency," using a rule- or learning-based method [22]–[25]. In addition, image processing leveraging multimodal learning has been an active research topic in the deep learning community in recent years (e.g., [26]). However, as Tsotsos notes [19], most of them do not fit the definition of active recognition because they do not aim to design a framework that contextually selects how, where, and when to recognize according to the purpose of the task, although human experiments clearly indicate the immense effects of top-down biases in recognition, owing to "selective attention" [27], [28]

Overall, our proposed verbal FoA is positioned as an extension of active recognition by leveraging the verbal instructions that co-occur with visual demonstration.

### C. Applications for symbol grounding

From the perspective of artificial intelligence, verbal FoA is considered a rudimentary form of "symbol grounding" [29], in the sense that verbal FoA aims to employ a physically situated natural language. Several works have been proposed to equip robots with not only symbol-grounding capabilities [30], [31], but also capabilities to compose grounded meanings [32]. Effective symbol grounding can solve a wide variety of problems in robotics-related applications, such as video object segmentation [33], vision-and-language navigation [34], and bidirectional mapping between human motions and natural languages [35]. Positioned in the same context as these studies, verbal FoA specifically aims to address the issue of spatio-temporally localizing human manipulation at the exact time of grasp and release.

## III. VERBAL FOA

In this section, we first explain the theoretical background of verbal FoA. Next, we describe detailed implementations to achieve verbal FoA. All modules were implemented on the robot operating system (ROS). Several examples of robot execution are also presented.

### A. FoA filters

The idea underlying verbal FoA is the "frame theory" developed by Marvin Minsky [36]. In theory, vision processing can solve a problem by confining the problem space using attention (i.e., *frame*), which depends on the situation. The core idea of this study is to apply the frames as a spatio-temporal filter to attenuate noise in an environment while simultaneously acquiring the necessary information to encode task models. In the frame theory, an appropriate set of frames should be prepared based on the problem to solve. In our case of "detecting timings and locations of grasping and releasing in the presence of spatio-temporal noises," a promising set of frames should include information regarding "where to grasp and release" and "when to grasp and release."

In the field of neuroscience, there is accumulating evidence that verbal instructions during demonstrations should contain useful information to support vision processing. For example, it has been shown that both visual information and verbal communication are used effectively in human-to-human demonstrations [37], [38]. Further, verbal information is known to affect visual attention and perceptual sensitivity in scenes with spatial uncertainty [12]–[14]. Therefore, verbal instructions should contribute to the formation of the "where" frame. Once the location is determined, the "when" frame can be computed by analyzing the interaction of a hand and an object in the "where" frame. In this way, the rationale for the verbally driven FoA can be provided.

Based on these considerations, we designed the verbal FoA to be composed of several sub-FoA filters:

- Target-object-location FoA: locations where the grasp and release occurred in a demonstration.
- Grasp-release FoA: timings when the grasp and release occurred in a demonstration.

In addition, we included three types of FoA to be used by a task-model encoder and the verbal FoA itself as additional frames:

- Target-name FoA: name of the target object.
- Attribute FoA: attributes of the target object.
- Task-candidate FoA: candidates of tasks to recognize.

### B. FoA working flow

Fig. 2 illustrates the pipeline used to calculate the verbal FoA. The verbal FoA is composed of several sub-FoA filters shown in the circles. The input data is transcribed verbal instructions, RGB-D images, and human skeleton poses during a demonstration. The modules shown in the boxes calculate the FoA filters using the following steps:

1. The language parser extracts task-related verbs (i.e., the task-candidate-FoA filter), a target-object name of the verbs (i.e., the target-name-FoA filter), and an attribute of the target object, such as color (i.e., the attribute-FoA filter).
2. The object selector calculates the time series of the target-object positions, which are represented in a voxel space (i.e., the target-object-location-FoA filter), by analyzing the target-name-FoA filter, the attribute-FoA filter, and all the detected object positions through a generic object detector.
3. Within the target-object locations, the grasp-release detector calculates the timings when the grasp and release occur with the laterality of a manipulating hand (i.e., the grasp-release-FoA filter), by analyzing the time-series distances of the target object to both hands.

In the current implementation of verbal FoA, we made several assumptions: 1) A demonstrated manipulation operation shall be associated with verbs, and 2) a target object shall not move spontaneously without human intervention.

### 1) Language parser

The language parser analyzes transcribed verbal instructions to output task-related verbs (i.e., the task-candidate-FoA filter), a target-object name (i.e., the target-name-FoA filter), and an attribute of the target object (i.e., the attribute-FoA filter). For example, if the input states to "Pick up a red cup and place it on the shelf.," the module outputs task-related verbs "[pick, place]," a target-object name "cup," and an attribute "red." The input data is obtained by sequentially applying an ego noise reduction filter [39], a voice activity detection based on signal power, and a cloud speech recognition service [40] to an audio signal recorded by a wireless microphone.

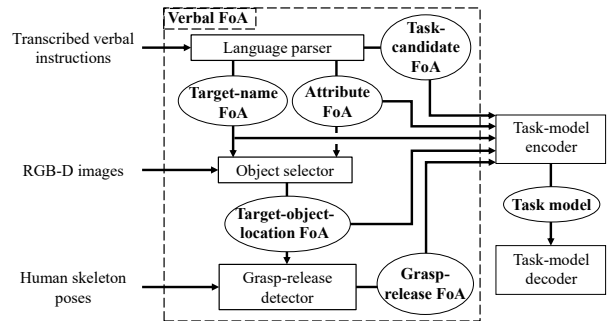

Figure 2. Overview of pipeline to produce verbal FoA (dashed rectangle), which is composed of several sub-FoA filters.

The module runs as follows. Firstly, it detects the word classes and the word dependencies using a Stanford language parser [41]. Next, the module filters task-related verbs by referring to a predefined set of task-related verbs, such as *lift*, *pick*, *put*, and *place*. This verb set was created by analyzing action verbs, which were spoken in YouTube instruction videos about daily chores [3]. Next, the module searches for a target object that is labeled as the accusative object of the filtered verbs. Finally, the module searches for adjectives that are labeled as adjectival modifiers of the object. The current implementation supports color as an attribute. Therefore, the module detects the color attribute, if any, by string-matching the adjectives that are found from the statement with a predefined color name set, which supports *red*, *yellow*, *green*, *cyan*, *blue*, *magenta*, *black*, and *white*.

### 2) Object selector

The object selector calculates the target-object-location-FoA filter by analyzing the target-name-FoA filter, attribute-FoA filter, and time-series RGB-D images. We used an Azure Kinect sensor [42] mounted on a robot head to capture the images. The nominal sampling rate was 5 Hz.

First, the module analyzes time-series RGB-D images to provide time-series data of the 3D positions of the detected objects as the outputs. To achieve this, a generic object detection CNN was used [43]. As we aim to teach household manipulation operations, we used an off-the-shelf third-party model [44], which detected 42 household objects such as *dish*, *cup*, and *plastic bottle* by taking an RGB image as an input. The module crops the RGB-D images using the detected bounding boxes (Fig. 3 (a), left pane) and converts them into point clouds represented by an environment coordinate. This conversion was calculated analytically using the intrinsic camera parameters. For each cropped image, the module calculates the object position as the mean value of the point cloud positions (Fig. 3 (a), right pane).

Next, this module filters objects by matching the names of the detected objects and the target-name-FoA filter (Fig. 3 (b), left pane). Following this, the module filters objects by matching object colors and the attribute-FoA filter (Fig. 3 (b), right pane). An object color is determined as the dominant pixel color inside the detected bounding box in the HSV color space. In the case where an attribute-FoA filter is not detected by the language parser, only the target-name-FoA filter is applied.

Finally, the filtered target objects are mapped to a 3D voxel space. In the current implementation, a voxel is a regular polygon of 0.15 m in length. Note that the module does not verify the identities between the target objects detected in different images. The target objects represented in the voxel space are sent to the grasp-release detector as the target-object-location-FoA filter.

*3) Grasp-release detector*

The grasp-release detector analyzes the human skeleton poses as well as the target-object-location-FoA filter to detect the timings when the grasp and release occur (Fig. 4). The human skeleton poses were obtained using an Azure Kinect sensor with a nominal sampling rate of 15 Hz. The positions labeled as *hand tips* by a driver API [42] were analyzed as human hand positions.

Assuming that the grasp and release occur when a human hand approaches and leaves a target object, the module first calculates the candidates of the timings using the equation:

$$T_i = Argmin\bigl(Distance(H_t, Obj_i)\bigr), \qquad (1)$$

where $i$ indicates the index of a 3D spatial voxel, $H_t$ indicates a hand position at time $t$, $Obj_i$ indicates the object position defined as the median of the target-object positions in a voxel $i$, $Argmin$ indicates an operation to obtain the index of the global minimum along the time axis, and $Distance$ indicates an operation to obtain the Euclidean distance between the two positions. $T_i$ is calculated for the left and right hands for all voxels containing more than one target object. We ignore the voxels where the distance between $H_t$ and $Obj_i$ at $T_i$ is more than 0.2 m, assuming that the grasp and release occur when the manipulating hand is close to the target object.

Next, the module decides whether a candidate pertaining to timing, $T_i$, corresponds to the grasp or release. To this end, the module analyses the averaged existence probabilities of the target object before and after $T_i$. Depending on the characteristics of the existence probabilities, the module classifies $T_i$ into one of the following three types: a) If the target object exists before $T_i$ but not after $T_i$, $T_i$ is classified as a grasp timing (Fig. 4 (b)) b) If the target object does not exist before $T_i$ but exists after $T_i$, $T_i$ is classified as a release timing (Fig. 4 (c)) c) Otherwise, $T_i$ is classified as an unrelated timing. In the current implementation, we set the criteria of existence probability as 0.5. Here, we assume that voxel resolution is sufficient that no grasp-release occurs in the same voxel. The $T_i$s that are classified as grasp or release are sent to a task-model encoder with the index $i$ of the voxels and the laterality of the manipulating hand.

*C. Use of verbal FoA to encode task models*

As the scope of this study is to detect timings and locations of grasping and releasing by verbal FoA, both the detailed design of task models, and algorithms to encode and decode them are beyond the scope of this study. However, to present the overview of the task-model encoder, this subsection shows a few examples of tasks and skill parameters defined herein as well as very brief explanations to compute them. The comprehensive design of the task models is explained in a separate paper [3].

Table 1 illustrates examples of tasks and skill parameters. We defined a task as the transition of contact states between a target object and an environment. Skill parameters are used to achieve state transitions by a robot. For example, we defined the *position parameters* and *grasp-release parameters* to control a robot by position control. Our LfO system predefines a set of a task and its associated empty skill parameters as a task-model template. When the system recognizes a task to be executed, the task template is loaded. Then, a process called daemon fills some of the empty skill parameters by analyzing the results of the verbal FoA. Some variant parameters, such as the grasp and release positions, are calculated at the moment of robot execution.

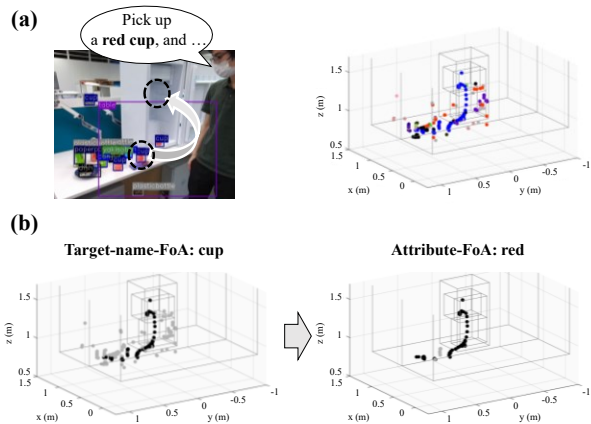

Figure 3. Processing of object selector. (a) (Left) Detected objects in an image. (Right) All detected object positions in demonstration overlaid on 3D environment model, which is depicted as gray lines. Each color depicts a different object class. (b) Object selection from all detected positions by target-name-FoA filter followed by attribute-FoA filter. Black points depict filtered positions by FoA filters. Gray points depict spatial noises.

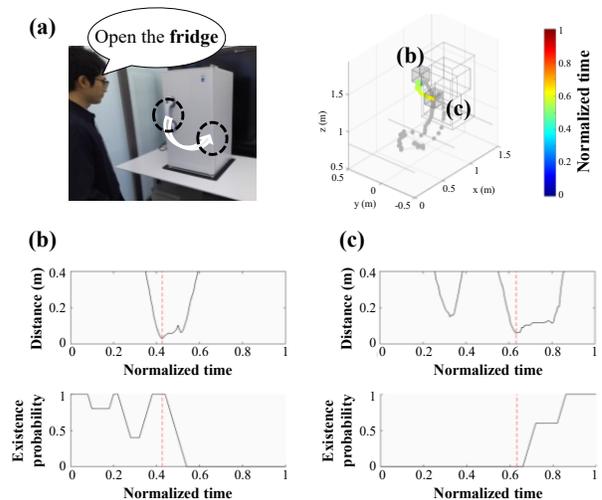

Figure 4. Processing of grasp-release detector. (a) (Left) Human demonstration of opening a fridge. (Right) Right-hand trajectory during demonstration. Colored points indicate hand positions filtered by timings of grasp and release. Gray points indicate temporal noises. (b, c) Analyses of the voxels where grasp and release were detected, respectively. Red dash lines indicate detected timings when grasp or release occurred. (Top) Euclidean distance between target object (i.e., fridge) and right hand. (Bottom) Existence probability of target object smoothed with time window corresponding to 10% of period of whole demonstration.

TABLE I  EXAMPLES OF TASKS AND SKILL PARAMETERS

| Task types | Position parameters | Grasp-release parameters |
|---|---|---|
| Pick-place an object | Waypoints | Object name and attribute, Grasp type, hand laterality, Grasp location, Release location, Grasp position, Release position |
| Rotate a linkage | Trajectory on a maintaining dimension | |

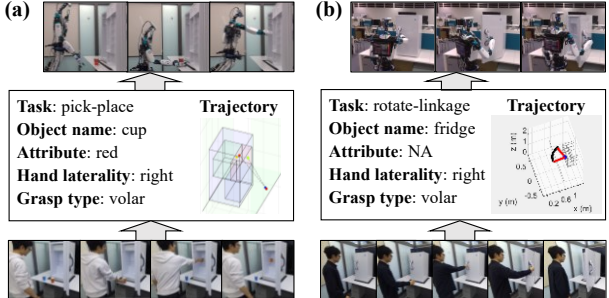

Figure 5. Examples of executions of an implemented LfO system. Substantiated task models with representative skill parameters are shown. (a) Teaching of pick-place task. (b) Teaching of rotate-a-linkage task.

Here, we explain a brief explanation of the computations of a task-model encoder. Task type is recognized by analyzing task-related verbs (i.e., task candidate-FoA filter) with a knowledge database associating verb sets with task types. For example, the verb *open* is associated with a rotate-a-linkage task. Position parameters are obtained by analyzing a manipulating-hand trajectory from the human skeleton poses between the grasp and release timings (i.e., the grasp-release-FoA filter). For the pick-place task, the position parameter is obtained as a spatially discretized hand trajectory. For the rotate-a-linkage task, the position parameter is obtained as parameters of a circle fitting to the hand trajectory. For the grasp-release parameters, object name, object attribute, and hand laterality are directly obtained from the verbal FoA. In addition, a daemon recognizes one of the human grasp types [45] at the moment when the grasp occurs (i.e., the grasp-release-FoA filter), using a pipeline that leverages an object affordance [46], [47]. The grasp and release locations are calculated as spatially discretized locations where the human grasped and released an object (i.e., the target-object-location-FoA filter). When executing a robot, the task-model decoder calculates the grasp and release positions inside the locations.

Fig. 5 shows examples of executions of an implemented LfO system. We used a humanoid robot, Seednoid [48], as the LfO agent. This robot was selected because it has a pair of 7-DOF arms as well as a movable waist to enable a wide variety of manipulations. The robot was controlled by position control using an IK solver [49]. A detailed explanation of robot execution is described in [50]. In the first case, a demonstrator picked up a red cup with the right hand and placed the cup on a shelf with a verbal instruction of "Pick up a red cup and place it on the shelf." The demonstration contained a blue cup as the spatial noise. In the second case, the demonstrator opened a fridge with the right hand with a verbal instruction of "Open the fridge." The demonstration contained unrelated body movements before and after the open-related movements. As a result of the first case, the system recognized the color attribute and the timings of grasp and release (Fig. 5 (a)). As a result of the second case, the system recognized the trajectory of the revolute movement by extracting the timings of grasp and release (Fig. 5 (b)). These results demonstrate examples of the verbal FoA operating combined with an LfO system.

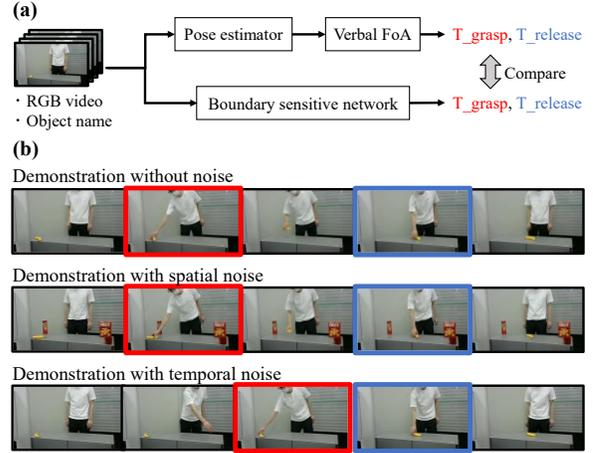

Figure 6. Overview of the experiment. (a) The schema of the comparison experiment. (b) Three types of demonstrations to test the methods. Red and blue rectangles depict the labeled grasp and release timings, respectively.

## IV. EVALUATION OF THE VERBAL FoA SYSTEM

We tested the robustness of the verbal FoA to spatio-temporal noises in a demonstration. We chose an end-to-end action localization network called "boundary sensitive network (BSN)" as a baseline method [51]. BSN outputs a pair of start and end time of an action by taking an RGB video as an input. We selected BSN as a baseline because it has shown one of the highest performances thus far in action localization tasks.

Fig. 6 (a) illustrates an overview. We compared the error of timing-estimation against ground-truth labels between the verbal FoA and BSN. To make the comparison fair, we used the same inputs: 1) RGB image sequences instead of RGB-D images and human skeleton poses, 2) a target-object name among ten household items in YCB objects (i.e., *apple*, *banana*, *chips can*, *cracker box*, *gelatin box*, *potted meat can*, *pitcher*, *bleach cleanser*, *bowl*, and *cup*) [52]. We used the YCB object set because it is a common benchmark for robotic manipulations, and it covers common household items. We simplified the verbal FoA to meet these inputs: 1) a third-party off-the-shelf human pose estimator [53] was used to detect hand positions in 2D space, 2) an input object name was treated as a target-name-FoA filter, 3) attribute FoA was ignored, and 4) target-object-location-FoA filter was calculated in 2D space, in which a grid size of ten pixels was used instead of a voxel. For BSN, the original network assumes the input to be a feature sequence calculated on the image sequence. To add information about an object name, a normalized value of the object-class index was added to each feature as a one-dimension feature.

A dataset of demonstrations without noise was recorded by five web-cameras (resolution of 640 × 360) from different positions simultaneously, with a nominal sampling rate of 30

Hz. A demonstrator performed a single pick-place task on each of the ten YCB objects eight times, each of which is a few-second video. The grasp and release locations were different each time. In total, 400 demonstration videos were obtained. A total of 376 videos were used to train the BSN network and to fine-tune an object detection network [44] for verbal FoA. The remaining 24 videos were used to test the methods. In addition, noisy demonstrations were recorded using the same recording setup for testing. As a demonstration with spatial noise, the pick-place task was performed with the existence of unrelated objects on the table (Fig. 6 (b)). As a demonstration with temporal noise, the task was performed with an unrelated "fake" motion before grasping (Fig. 6 (b)). We recorded 24 videos for each type of noisy demonstration, in which the object, camera location, and grasp and release locations were randomized. Ground-truth timings of the grasp and release were labeled manually. In total, 72 videos were tested under three conditions.

While carrying out testing, we observed some cases where the verbal FoA failed to detect either grasp or release. In most of the cases, the grasp-release detector miscalculated $T_i$ due to the difficulty in measuring the exact distance between a hand and an object in 2D space. The difficulty was specific to the experimental setup, where the verbal FoA was simplified. In a few cases, the generic object detector failed to detect a target object in some parts of the image sequence. As we did not attribute both types of failures directly to the spatio-temporal noises, these cases were excluded from the comparison in order to focus on a comparison of robustness against noise. Fig. 7 shows the error in the timing-estimation of the two methods. While estimations by BSN were compromised by temporal noise, the estimation by verbal FoA was less affected by both types of noise, indicating the effectiveness of the verbal FoA in terms of robustness to spatio-temporal noises.

## V. Discussion

### A. Role of FoA as a guide of the vision system

The comparison experiment revealed the robustness of the verbal FoA against an end-to-end system. The idea of verbal FoA is to confine the problem space that a vision system solves using additional knowledge (e.g., a name of the target object). In that sense, our system is an implementation of a form of the "frame theory," by Marvin Minsky [36]. Although this study highlights the aspect of FoA (i.e., *frame*) as a noise filter, the defined FoA filters could also contribute to providing a task-model-encoder with skill parameters (see Table 1 and Fig. 5).

Temporal noise affected the estimations by BSN in such a way that they shifted to earlier times (Fig.7 bottom-right pane). The method as well as the majority of other action recognition networks relies on temporal information, such as optical flow; the estimations might have been compromised by the "fake" motions. Precise detection of grasping and releasing can be critical to obtaining skill parameters. For example, it has been suggested [54] that a grasp posture begins to converge after 80% of the reaching period, suggesting that a grasp posture could not be determined until just before the grasping time. Therefore, the shift due to the temporal noise can be fatal to grasp-type recognition.

While the verbal FoA showed its robustness, we observed a few cases (3/74 videos) where the object detector failed to detect a target object in some parts of the image sequence. As the system relies on accurate estimation of the object, verbal instruction, and human poses, guaranteeing these performances is a key issue. There are two possible approaches. First, as in a human-to-human demonstration, the system could equip with a questioning module to compensate for any lacking information. The role of verbal communication for disambiguation is also supported by LfD research [55]. Another solution is to use the help of a remote human operator under a paradigm such as the "human-robot cloud" [56]. These solutions would contribute not only to the rapid deployment of the LfO system in the real world but also to creating empirically derived datasets that would further catalyze the increasing autonomy of the method.

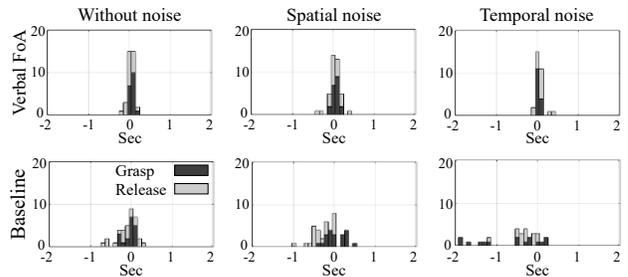

Figure 7. Errors in the estimation of timings of grasp and release. Stacked histograms are shown as time difference from the ground-truth labels.

### B. Toward a better symbol grounding

In this study, we used a knowledge database associating verb sets with task types to detect a task. However, the verb-based association may be insufficient because the association could change depending on the affordance of the target object [56]. For example, while "open a door" suggests a rotate-a-linkage task, "open a drawer" suggests a sliding task. This issue was recently addressed by Paulius et al. [57]. Thus, the database can be extended by including the specific affordances of each object. In addition, considering that attributes contribute to ensuring the identity and homogeneity of the target object, the system can more efficiently guide the task-model encoder by supporting other attributes-FoA filters beyond color. To the best of our knowledge, there is no agreement as to a practically sufficient set of attributes. As a potential baseline, Beetz et al. proposed the use of color, size, position, and shape for a robot perception system operating in a household environment [58]. An extension of this study would need to cover these attributes.

## VI. Conclusion

In this study, in the context of LfO, we dealt with the problem of detecting timings and locations of grasping and releasing in the presence of spatio-temporal noises. Inspired by the frame theory and a function of verbal instructions to guide an observer's visual attention, we proposed a verbal FoA system to reduce the noise, specifically caused by unrelated objects in an environment and human body movements before and after a manipulation operation. We demonstrated the robustness of the verbal FoA in attenuating the noise by comparing it with a state-of-the-art vision system for action localization. The results suggest the usefulness of the verbal FoA for the deployment of an LfO system in the real world.